\documentclass[letterpaper, 10pt, conference]{ieeeconf}      % Use this line for a4 paper

\IEEEoverridecommandlockouts                              % This command is only needed if 
\usepackage{cite}
\usepackage{color}
\usepackage[dvipsnames]{xcolor}
\usepackage{amsmath}

\usepackage{epsfig}
\usepackage{amssymb}

\title{\LARGE \bf
Sensitivity of Legged Balance Control\\ to Uncertainties and Sampling Period
}

\author{Nahuel A. Villa$^{1}$, Johannes Englsberger$^{2}$ and Pierre-Brice Wieber$^{1}$% <-this % stops a space
\thanks{This work was funded by the EU H2020 Comanoid Research and Innovation Action (RIA).}% <-this % stops a space
\thanks{$^{1}$Univ. Grenoble Alpes, Inria, 38000 Grenoble, France 
        {\tt\small nahuel.villa@inria.fr;\newline\tt\small pierre-brice.wieber@inria.fr.}}%
\thanks{$^{2}$German Aerospace Center (DLR), Institute of Robotics and Mechatronics, 82234 Wessling, Germany.
        {\tt\small johannes.englsberger@dlr.de}}%
}

\begin{document}

\maketitle
\thispagestyle{empty}
\pagestyle{empty}

%%%%%%%%%%%%%%%%%%%%%%%%%%%%%%%%%%%%%%%%%%%%%%%%%%%%%%%%%%%%%%%%%%%%%%%%%%%%%%%%
\begin{abstract}
	
We propose to quantify the effect of sensor and actuator uncertainties on the control of the center of mass and center of pressure in legged robots, since this is central for maintaining their balance with a limited support polygon. Our approach is based on robust control theory, considering uncertainties that can take any value between specified bounds. This provides a principled approach to deciding optimal feedback gains. Surprisingly, our main observation is that the sampling period can be as long as $200$~ms with literally no impact on maximum tracking error and, as a result, on the guarantee that balance can be maintained safely. Our findings are validated in simulations and experiments with the torque-controlled humanoid robot Toro developed at DLR. The proposed mathematical derivations and results apply nevertheless equally to biped and quadruped robots.

\end{abstract}

%%%%%%%%%%%%%%%%%%%%%%%%%%%%%%%%%%%%%%%%%%%%%%%%%%%%%%%%%%%%%%%%%%%%%%%%%%%%%%%%
\section{Introduction}

%%%%%%%%%%%%%%%%%%%%

Biped and quadruped robots are beginning now to master the skill of walking dynamically in most standard situations~\cite{Fengpaper, Mastalli2017QuadrupedRoughTerrain, Caron2018stairClimbing}. This suggests that more widespread commercial use of such robots will soon be possible. This requires, however, that guarantees are provided about their safety and operational performance. In research prototypes, the risk of failure is usually contained by using very fast and precise (and therefore very expensive) sensors, actuators and computers, resulting in robots that are clearly too expensive for commercial purposes.

The dynamics of the Center of Mass (CoM) of these robots over the support feet is unstable, and therefore very sensitive to all sources of uncertainties. But how fast and precise, and therefore how expensive should the sensors, actuators and computers be has never been investigated in the existing scientific literature. A precise quantification of the effect of uncertainties and sampling period on legged balance control seems to be missing, and it is the goal of this paper to initiate this discussion.

The balance of legged robots mostly involves motion of their CoM with respect to their feet on the ground. We therefore focus our analysis on the motion of the CoM, considering that other aspects of the motion of the robot, such as precise whole-body joint motion and contact force control, are handled separately, as usual in this field of robotics~\cite{Handbook}.

We introduced in~\cite{Villa2017} a tube-based Model Predictive Control (MPC) of walking in order to guarantee that all kinematic and dynamic constraints are always satisfied, even in the presence of uncertainties. We considered that uncertainties can take any value between some bounds, generating some tracking error which can be bounded accordingly. Here, we propose to analyse how these bounds are related: how much tracking error can we expect for a given amount of uncertainty? This naturally depends not only on the kind of uncertainty (e.g. on sensors or actuators), but also on the control law and its sampling period.

\begin{figure}
\centering	
\includegraphics[width=\linewidth]{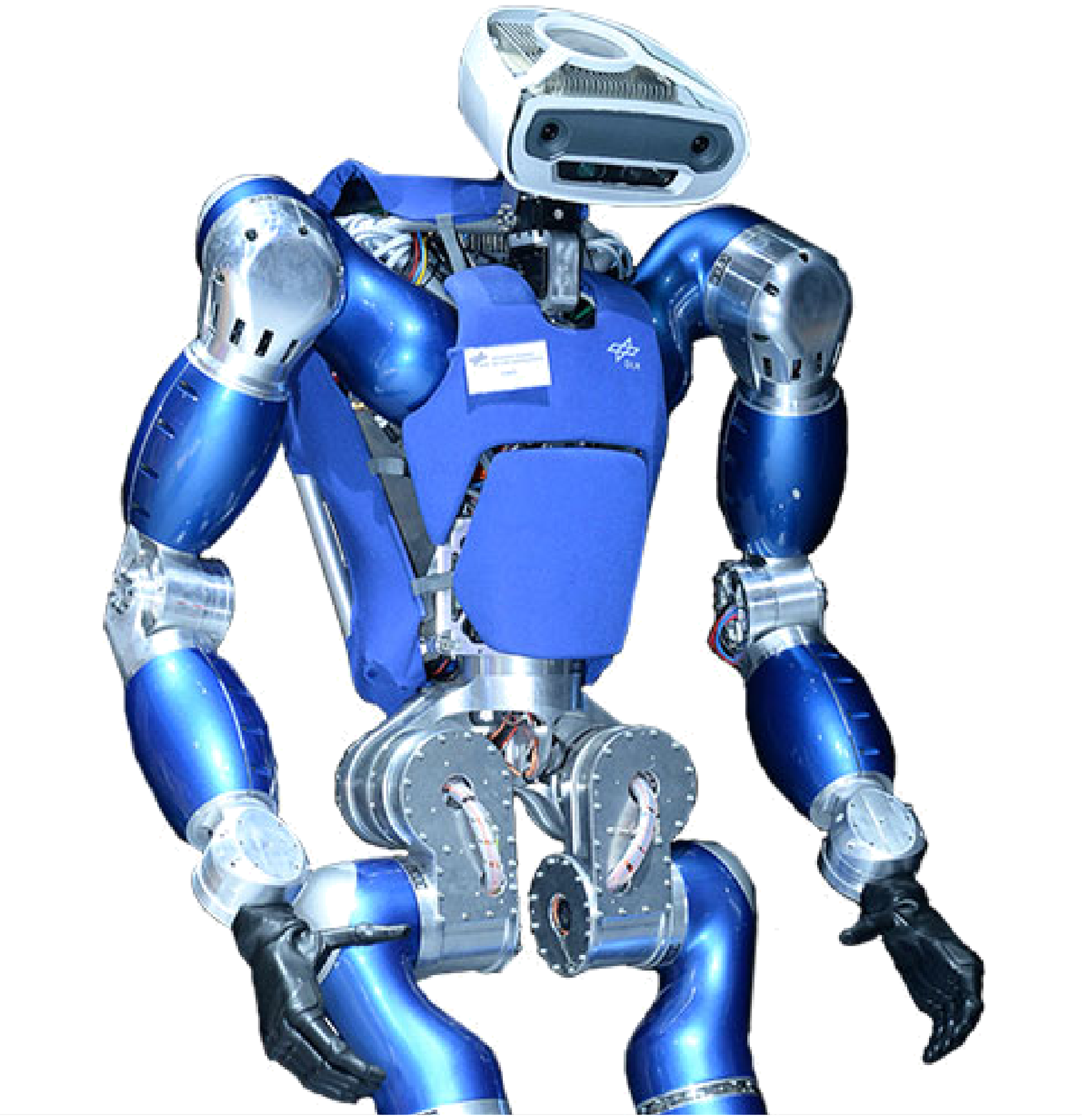}
\caption{Toro is a torque-controlled humanoid robot developed at DLR~\cite{Johannes2018filterVibra}.} \label{The robot}
\end{figure}

Our findings are validated in experiments and simulations with the torque-controlled humanoid robot Toro developed at DLR (Fig.~\ref{The robot}). The proposed mathematical derivations and results apply nevertheless indistinctly to biped and quadruped robots.

Section~\ref{Biped Walking Modelling} introduces basic aspects of the CoM dynamics. A standard feedback control law is proposed and conditions for its stability are determined in Section~\ref{App: Stability Limits}. The dynamics of the tracking error is analysed in Section~\ref{Tracking control} and related to the bounded uncertainties in Section~\ref{Sec: CompRatio}. Feedback gains are then optimized to minimize the span of the tracking error in Section~\ref{Sec: OptimalGains}. Our theoretical analysis is validated experimentally and in simulations on the torque-controlled humanoid robot Toro in Section~\ref{Sec: Numerical Example}. Finally, we summarize our conclusions in Section~\ref{Sec: Conclusions}.

\section{Walking Model} \label{Biped Walking Modelling}

Consider a legged robot walking on flat, horizontal ground. The Center of Pressure (CoP) $p$ of the contact forces with the ground can be related to the motion of the Center of Mass (CoM) $c$ of the robot and its angular momentum $L$ as follows~\cite{Handbook}:
\begin{equation} \label{eq. SystemDynamics}
p^{x,y}=c^{x,y}-\frac{mc^{z}\ddot{c}^{x,y}-S\dot{L}^{x,y}}{m(\ddot{c}^{z}+g^{z})},
\end{equation}
where $^x$ and $^y$ indicate horizontal coordinates, $g^z$ is the vertical acceleration due to gravity, $m$ the mass of the robot and $S=\left[\begin{smallmatrix}0&-1\\1&0\end{smallmatrix}\right]$ a $\frac{\pi}{2}$ rotation matrix. Due to the unilaterality of contact forces, this CoP is bound to the support polygon $\mathcal{P}(t)$, which varies with time depending on which feet are in contact with the ground and where:
\begin{align} \label{eq. CoP RealConstraint}
p&\in\mathcal{P}(t).
\end{align}

This can be reformulated as a dynamics
\begin{equation}
\ddot{c}^{x,y}=\omega^{2}(c^{x,y}-p^{x,y}+n^{x,y}),
\end{equation}
with some constant value $\omega^{2}\approx\frac{g^z}{c^z}$, gathering all non-linearities in a vector
\begin{equation}
n=\frac{\ddot{c}}{\omega^2}-\frac{mc^z\ddot{c}-S\dot{L}}{m(\ddot{c}^z+g^z)},
\end{equation} 
which can be bounded efficiently~\cite{Camille, Serra2016newton}:
\begin{equation}
n\in\mathcal{N}.
\end{equation}
Since the $^x$ and $^y$ coordinates appear to be decoupled, we will consider only the $^x$ coordinate in the following.

Assuming that $p$ and $n$ are constant over time intervals of length $\tau$, we can obtain a discrete-time linear system following a standard procedure~\cite{Ogata}:
\begin{equation} \label{DiscretSystem}
x^{+}=Ax+B(p-n),
\end{equation}
with matrices
\begin{align}
A&=\begin{bmatrix}\cosh(\omega\tau)&\omega^{-1}\sinh(\omega\tau)\\
\omega\sinh(\omega\tau)&\cosh(\omega\tau)\end{bmatrix} \label{MatrixA},\\
B&=\begin{bmatrix}1-\cosh(\omega\tau)\\
-\omega\sinh(\omega\tau)\end{bmatrix} \label{MatrixB},
\end{align}
and $x^+$ the successor of the state
\begin{equation} \label{CoM RealConstraint}
x=\begin{bmatrix}c^x\\\dot{c}^x\end{bmatrix}\in\mathcal{X}(t),
\end{equation}
where $\mathcal{X}(t)$ represents time-varying kinematic constraints on the CoM motion.

\section{Stable Feedback Gains} \label{App: Stability Limits}

We control the CoP $p$ using a linear feedback with compensation of $n$:
\begin{equation} \label{TrackingControlLaw1}
p=p_\mathit{ref}+K(x-x_\mathit{ref})+n
\end{equation}
with feedback gains of the form
\begin{equation} \label{Kform}
K=k\begin{bmatrix}1&\lambda\end{bmatrix}
\end{equation}
in order to track a reference trajectory $x_\mathit{ref}$, $p_\mathit{ref}$ (obtained with any standard motion generation scheme~\cite{Handbook}). If the reference trajectory follows the dynamics~\eqref{DiscretSystem} without $n$:
\begin{equation}
x^{+}_\mathit{ref}=Ax_\mathit{ref}+Bp_\mathit{ref},
\end{equation}
this leads to a closed-loop dynamics
\begin{equation}
\tilde{x}^+=(A+BK)\tilde{x} \label{TrErrorClosed-loop Dynamics1}
\end{equation}
of the tracking error
\begin{equation} \label{TrErrorState}
\tilde{x}=x-x_\mathit{ref}.
\end{equation}

\begin{figure}
\centering
\hspace*{-7.0mm}\includegraphics[width=1.13\linewidth]{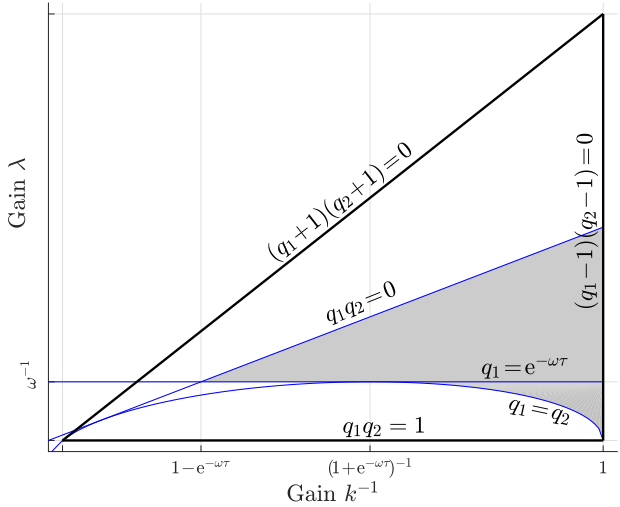}
\caption{Following Jury's simplified stability criterion, the region of feedback gains $k^{-1}$ and $\lambda$ that lead to a stable closed-loop dynamics is a triangle, where $\lambda$ varies between $\frac{\cosh(\omega\tau)-1}{\omega\sinh(\omega\tau)}$ and $\frac{\cosh(\omega\tau)+1}{\omega\sinh(\omega\tau)}$ while $k^{-1}$ varies between $\frac{\cosh(\omega\tau)-1}{\cosh(\omega\tau)+1}$ and $1$. The thin blue lines correspond to having both poles equal, at least one pole equal to $\mathrm{e}^{-\omega\tau}$ (when $\lambda=\omega^{-1}$), or at least one pole equal to zero. The gray area corresponds to having both poles positive real, and at least one greater or equal to $\mathrm{e}^{-\omega\tau}$.} \label{StabilityAreaWithLines}
\end{figure}

Consider the two poles, $q_1$ and $q_2$, related to this closed-loop dynamics as follows:
\begin{align}
q_1q_2&=\det(A+BK)\nonumber\\
&=1-k+k\cosh(\omega\tau)-k\lambda\omega\sinh(\omega\tau),\label{Determinant}\\
q_1+q_2&=\mathrm{tr}(A+BK)\nonumber\\
&=k+(2-k)\cosh(\omega\tau)-k\lambda\omega\sinh(\omega\tau).\label{Trace}
\end{align}
Following Jury's simplified stability criterion~\cite{Jury1962simplified}, this closed-loop dynamics is stable if and only if:
\begin{align}
q_1q_2&<1\label{App: Det=0},\\
(q_1-1)(q_2-1)=q_1q_2-(q_1+q_2)+1&>0,\label{StabilityForRealPoles}\\
(q_1+1)(q_2+1)=q_1q_2+(q_1+q_2)+1&>0,
\end{align}
which corresponds to the following constraints represented in Fig.~\ref{StabilityAreaWithLines}:
\begin{align}
\lambda&>\frac{\cosh(\omega\tau)-1}{\omega\sinh(\omega\tau)}, \label{eq. lam lower bound}\\
k&>1,\label{eq. k lower bound}\\
k\lambda&<\frac{\cosh(\omega\tau)+1}{\omega\sinh(\omega\tau)}.
\end{align}

\section{From Uncertainties to Tracking Error}\label{Tracking control}

Consider that the CoP is affected by a bounded additive uncertainty
\begin{equation}
v\in W \label{v belongs W}
\end{equation}
coming from actuators, sensors and model errors (a precise expression will be
introduced in Section~\ref{Sec: OptimalGains}), such that the CoP in the linear
feedback~\eqref{TrackingControlLaw1} actually is:
\begin{equation} \label{TrackingControlLaw}
p=p_\mathit{ref}+K(x-x_\mathit{ref})+n+v.
\end{equation}
The closed-loop dynamics~\eqref{TrErrorClosed-loop Dynamics1} becomes
\begin{equation}
\tilde{x}^+=(A+BK)\tilde{x}+Bv. \label{TrErrorClosed-loop Dynamics}
\end{equation}

If the closed-loop matrix $A+BK$ is stable, then when time tends to infinity, the tracking error $\tilde{x}$ converges to a set $Z$:
\begin{equation}
\tilde{x}\rightarrow Z=\bigoplus_{i=0}^\infty(A+BK)^iBW,			\label{TheInvariantSet}
\end{equation}
where the symbol $\oplus$ represents a Minkowski sum\footnote{Given sets $A$ and $B$, $A\oplus B=\{a+b \mid a\in A,b\in B\}.$}. Following \eqref{TrErrorClosed-loop Dynamics} and \eqref{TheInvariantSet}, once the tracking error is in $Z$, it stays in $Z$ for every future time \cite{Seron}:
\begin{equation}
\forall v\in W ,\ \tilde{x}\in Z\implies\tilde{x}^+\in Z.
\end{equation}
We use this robust positive invariance property to ensure a bounded tracking error
\begin{equation}
\tilde{x}\in Z,
\end{equation}
provided that the robot motion starts within these bounds. As an example, the
tube-based MPC scheme proposed in~\cite{Villa2017} for biped walking generates
the reference motion $x_\mathit{ref}$ online under this condition.

This precise bound on the tracking error allows guaranteeing that the
kinematic constraint~\eqref{CoM RealConstraint} will always be satisfied, even
with the uncertainty~\eqref{v belongs W}, provided that
\begin{equation}
x_\mathit{ref}\in\mathcal{X}(t)\ominus Z, \label{ConstrainedTracking error State}
\end{equation}
where the symbol $\ominus$ represents a Pontryagin difference\footnote{Given
sets $A$ and $B$, $A\ominus B=\{x\mid x+B\subseteq A\}.$}. In this case, the
corresponding CoP tracking error 
\begin{equation}
\tilde{p}=p-p_\mathit{ref}-n=K\tilde{x}+v
\end{equation}
is bounded accordingly:
\begin{equation}
\tilde{p}\in KZ\oplus W, \label{CoPTracking error Bound}
\end{equation}
so we can guarantee that the support polygon constraint~\eqref{eq. CoP
RealConstraint} will be satisfied as well, provided that
\begin{equation}
p_\mathit{ref}\in\mathcal{P}(t)\ominus\mathcal{N}\ominus KZ\ominus W. \label{ConstrainedTracking error CoP}
\end{equation}

Feasibility of the reference motion generation and tracking imposes that the
sets in~\eqref{ConstrainedTracking error State} and~\eqref{ConstrainedTracking
error CoP} are non-empty. The support polygon $\mathcal{P}(t)$ constraining the
CoP is normally smaller than the kinematic constraints $\mathcal{X}(t)$ on the
CoM motion. And the bound $KZ\oplus W$ on the CoP tracking error is larger than
the bound $Z$ on the CoM tracking error when using stable gains $K$, satisfying
condition~\eqref{eq. k lower bound}. Thus, as usual in the balance of legged
robots, the constraint~\eqref{ConstrainedTracking error CoP} on the CoP is the
limiting factor, and we look to reduce specifically the bound $KZ\oplus W$ on
the CoP tracking error.

\section{CoP Tracking Error due to Uncertainties} \label{Sec: CompRatio}

Using definition~\eqref{TheInvariantSet} of the set $Z$, the
bound~\eqref{CoPTracking error Bound} on the CoP tracking error becomes:
\begin{equation}
\tilde{p}\in\bigoplus_{i=0}^\infty K(A+BK)^iBW\oplus W.
\end{equation}
Considering a real interval
\begin{equation}
W=[v_\mathit{min},v_\mathit{max}],
\end{equation}
the maximum and minimum values for $\tilde{p}$ are reached with opposite
sequences of maximum and minimum values $v_\mathit{max}$ and $v_\mathit{min}$,
depending on the sign of each real coefficient $K(A+BK)^iB$. This results in
\begin{equation}
\tilde{p}_\mathit{max}-\tilde{p}_\mathit{min}=\left(\sum_{i=0}^\infty\left|K(A+BK)^iB\right|+1\right)(v_\mathit{max}-v_\mathit{min}).
\end{equation}
We can introduce then the spans
\begin{equation}
\tilde{p}_\mathit{span}=\tilde{p}_\mathit{max}-\tilde{p}_\mathit{min}
\end{equation}
and
\begin{equation}
v_\mathit{span}=v_\mathit{max}-v_\mathit{min},
\end{equation}
and the ratio
\begin{equation} \label{Amplification}
r=\frac{\tilde{p}_\mathit{span}}{v_\mathit{span}}=\sum_{i=0}^\infty\left|K(A+BK)^iB\right|+1
\end{equation}
between the amount of uncertainty and the resulting amount of CoP tracking
error.

The gray area in Fig.~\ref{StabilityAreaWithLines} corresponds to having both
poles $q_1$ and $q_2$ positive real, and at least one greater or equal to
$\mathrm{e}^{-\omega\tau}$. The extent of this area depends on the product
$\omega\tau$, but inside this area, the above ratio is
\begin{equation}
r=\frac{1}{k-1}+2,
\end{equation}
as shown in the Appendix, which is surprisingly independent from $\lambda$,
$\omega$ and $\tau$. It can be observed numerically that this is actually the
minimum possible ratio. Within this area, the choice $\lambda=\omega^{-1}$ is
particularly interesting since it maximizes controllability~\cite{Best}. We
consider therefore feedback gains of the form:
\begin{equation} \label{Ksugihara}
K=k\begin{bmatrix}1&\omega^{-1}\end{bmatrix}.
\end{equation}

In that case, we obtain poles
\begin{align}
q_1&=\mathrm{e}^{-\omega\tau},\\
q_2&=1-(k-1)(\mathrm{e}^{\omega\tau}-1)
\end{align}
from~\eqref{Determinant} and~\eqref{Trace}. Borrowing from the Appendix the
reformulation~\eqref{infSum} of the infinite sum~\eqref{Amplification} with
coefficients~\eqref{eq. Alpha1} and~\eqref{eq. Alpha2}, we obtain that the above
ratio becomes
\begin{equation}
r=\frac{k(\mathrm{e}^{\omega\tau}-1)}{1-|q_2|}+1.
\end{equation}
Depending on the sign of $q_2$ (positive being inside of the gray area, negative
being outside), we have:
\begin{equation} \label{CR -- Z-CC -- Sugihara line}
r=\begin{cases}
\frac{1}{k-1}+2&\text{if}\ \mathrm{e}^{\omega\tau}-1\leq\frac{1}{k-1},\vspace{2ex}\\
\frac{2+(\mathrm{e}^{\omega\tau}-1)}{2-(k-1)(\mathrm{e}^{\omega\tau}-1)}&\text{if}\ \frac{1}{k-1}\leq\mathrm{e}^{\omega\tau}-1<\frac{2}{k-1}.
\end{cases}
\end{equation}
When $\mathrm{e}^{\omega\tau}-1\geq\frac{2}{k-1}$, the closed loop is unstable and the ratio $r$ is undefined.

\section{Optimal Gains and Sampling Periods} \label{Sec: OptimalGains}

With feedback gains of the form~\eqref{Ksugihara}, the linear feedback with
uncertainties~\eqref{TrackingControlLaw} can be reformulated as
\begin{equation} \label{**}
p=p_\mathit{ref}+k(\xi-\xi_\mathit{ref})+n+v,
\end{equation}
where
\begin{equation}
\xi=c+\omega^{-1}\dot{c}
\end{equation}
is the Capture Point (CP)~\cite{Handbook}. Considering an error $\hat{\xi}$ in
the estimation of the CP $\xi$ and an error $\hat{n}$ in the model of the robot
(including inaccuracies in the actuation and ground contact), this linear
feedback actually becomes
\begin{equation}
p=p_\mathit{ref}+k(\xi+\hat{\xi}-\xi_\mathit{ref})+n+\hat{n},
\end{equation}
corresponding to an uncertainty $v$ of the form:
\begin{equation}
v=k\hat{\xi}+\hat{n}.
\end{equation}

Using the ratio~\eqref{Amplification}, the resulting span of CoP tracking error is:
\begin{equation} \label{Combined}
\tilde{p}_\mathit{span}=rk\,\hat{\xi}_\mathit{span}+r\,\hat{n}_\mathit{span}.
\end{equation}
Based on~\eqref{CR -- Z-CC -- Sugihara line}, its minimum value 
\begin{equation}\label{eq. minimumTrackingError}
\tilde{p}_\mathit{span}^*=\left(\sqrt{\hat{\xi}_\mathit{span}} + \sqrt{2(\hat{\xi}_\mathit{span}+\hat{n}_\mathit{span})}\right)^{2}
\end{equation}
is obtained using a feedback gain
\begin{equation}\label{eq. q. optimalGain}
k^* = 1+\sqrt{\dfrac{\hat{\xi}_\mathit{span}+\hat{n}_\mathit{span}}{2\hat{\xi}_\mathit{span}}}.
\end{equation}
Typical values for these sources of uncertainties are
$\hat{n}_\mathit{span}=\hat{\xi}_\mathit{span}=1$~cm
\cite{Flayols2017experimental}, resulting in a minimal span of CoP tracking
error $\tilde{p}_\mathit{span}^*=9$~cm, corresponding to Toro's feet width. In
this case, the optimal gain is $k^*=2$.

\begin{figure}
\centering
\hspace*{-4.3mm}\includegraphics[width=1.13\linewidth]{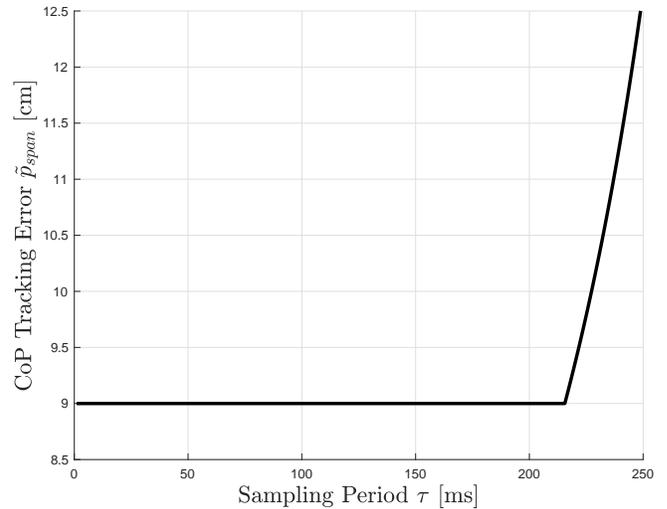}%soloUna(Z-CC)T
\caption{Span of the CoP tracking error $\tilde{p}_\mathit{span}$ produced by model and estimation errors with span $\hat{n}_\mathit{span}=\hat{\xi}_\mathit{span}=1$~cm, using the optimal feedback gains $k=2$ and $\lambda=\omega^{-1}$ ($\omega\approx3.2$~s$^{-1}$ for Toro) for different sampling periods $\tau$. The tracking error degrades sharply when $\tau>\omega^{-1}\ln2=216$~ms, but it doesn't improve for sampling periods below this value.} \label{CombinedT}
\end{figure}

But the key observation in~\eqref{CR -- Z-CC -- Sugihara line} is that once a gain $k$ has been decided, the ratios $r$ and $rk$ don't depend on the sampling period $\tau$, as long as it is shorter than
\begin{equation} \label{UmbralT}
\tau_0=\omega^{-1}\ln\left(\frac{1}{k-1}+1\right).
\end{equation}
The maximum CoP tracking error $\tilde{p}_\mathit{span}$ is not improved by reducing the sampling period below this value, but it degrades sharply when $\tau>\tau_0$, as shown in Fig.~\ref{CombinedT}. When $k=2$, $\tau_0=\omega^{-1}\ln2=216$~ms ($\omega\approx3.2$~s$^{-1}$ for Toro).

\section{Experimental Results} \label{Sec: Numerical Example}

The CP linear feedback~\eqref{**} is implemented in the humanoid robot Toro, with a feedback gain $k=2$ as discussed above, together with a standard Quadratic Program (QP) based inverse dynamics scheme for Whole-Body Control (WBC) of joint positions and contact forces~\cite{Johannes2018filterVibra}. The sampling period of the QP-based WBC is kept constant at $3$~ms while varying the sampling period $\tau$ of the CP feedback~\eqref{**}. The reference trajectory for a simple sequence of steps is actually not adapted to the sampling period, making it more difficult to track precisely at each contact transition with longer sampling periods (see video). 

\begin{figure}
\centering
\hspace*{-4.5mm}\includegraphics[width=1.13\linewidth]{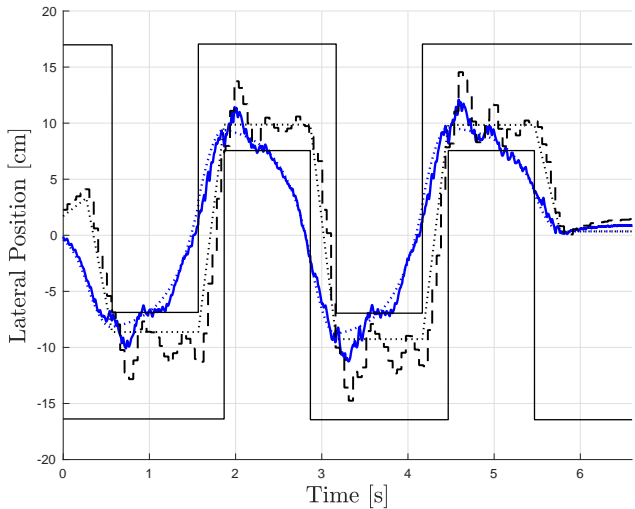}
\hspace*{-4.5mm}\includegraphics[width=1.13\linewidth]{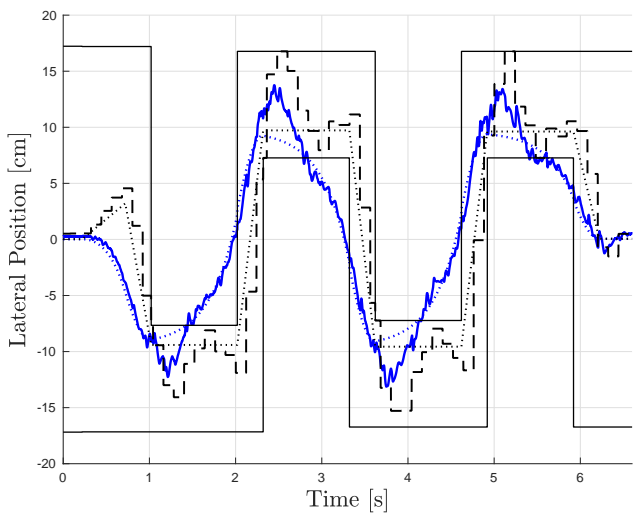}
\caption{Lateral component of walking experiments with the humanoid robot Toro using a feedback gain $k=2$ and sampling period $\tau=51$~ms (top) or $\tau=120$~ms (bottom). The CP $\xi$ is represented in blue, while the CoP is in dashed black. The reference values $\xi_\mathit{ref}$ and $p_\mathit{ref}$ are indicated with dotted lines.} \label{walkingExp2}
\end{figure}

\begin{figure}
\centering
\hspace*{-5mm}\includegraphics[width=1.13\linewidth]{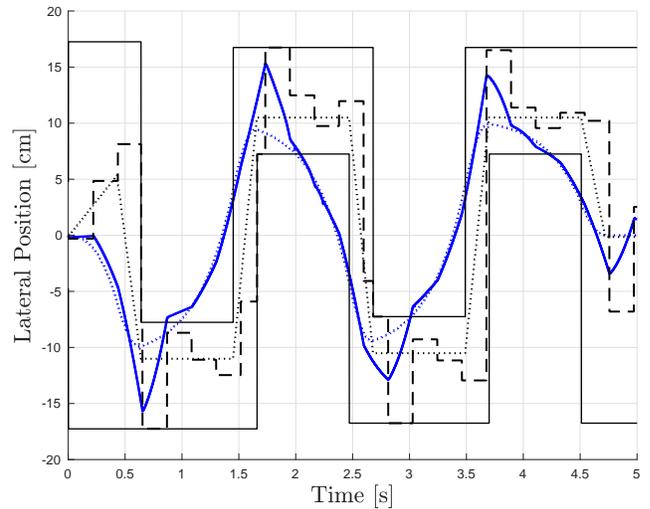}
\hspace*{-5mm}\includegraphics[width=1.13\linewidth]{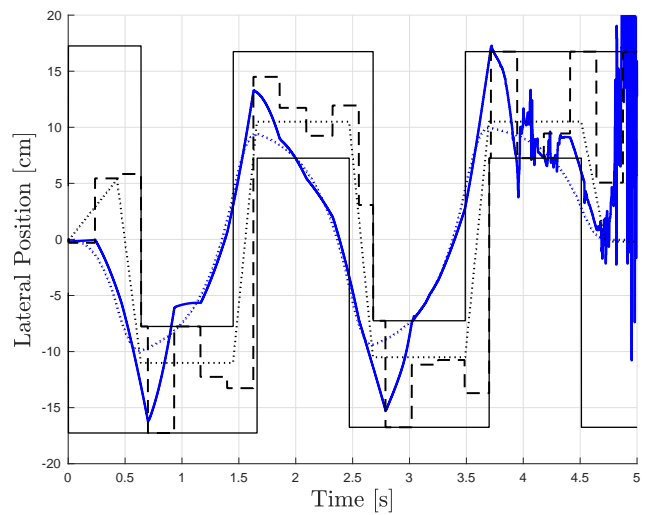}
\caption{Lateral component of walking simulations with the humanoid robot Toro using a feedback gain $k=2$ and sampling periods $\tau=216$~ms (top) or $\tau=232$~ms (bottom). The CP $\xi$ is represented in blue, while the CoP is in dashed black. The reference values $\xi_\mathit{ref}$ and $p_\mathit{ref}$ are indicated with dotted lines.} \label{walkingSim}
\end{figure}

We can observe in Fig.~\ref{walkingExp2} that in experiments with Toro, the lateral CP and CoP tracking performances are similar and satisfactory when $\tau=51$~ms or $120$~ms, as expected from our theoretical analysis. For longer sampling periods, the WBC generates larger arm motions in order to compensate angular momentum variations, which ends up triggering an emergency stop due to the increased risk of collision (see video). The resulting failure originates in the QP-based WBC and not the CP linear feedback~\eqref{**}, so this doesn't contradict the proposed theoretical analysis. In simulations, this safety system is not triggered and we can observe in Fig.~\ref{walkingSim} that the tracking performance is maintained at a satisfactory level for sampling periods up to $\tau=216$~ms while degrading sharply afterwards, validating strikingly well the theoretical analysis proposed above.

\section{Discussion and Conclusion} \label{Sec: Conclusions}

We quantify the effect of sensor and actuator uncertainties on the CoM and CoP tracking error in legged robots, since this is central for maintaining their balance with a limited support polygon. Our approach is based on robust control theory, considering uncertainties that can take any value between some bounds. The relationships we obtain can be used during the design stage of a legged robot, when looking for the best compromise between sensor, actuator, and CPU performance and cost. This principled approach also provides the corresponding optimal feedback gains.

Our main observation is that the sampling period for a human-sized humanoid robot such as Toro can be as long as $200$~ms with literally no impact on maximum tracking error and, as a result, on the guarantee that balance can be maintained safely. Concerning quadruped robots, stable locomotion has been realized recently with similarly low, $15$~Hz control rates~\cite{Grandia2019feedback-MPC-ETH}. Faster sampling periods might be useful for other aspects of the motion of the robot, such as arm or swing leg motion, but not for CoM motion.

This provides some freedom in the choice of the sampling period, which helped us achieve a substantial reduction of the oscillations mentioned in~\cite{Johannes2018filterVibra} by avoiding structure resonance modes. This could also help reduce energy consumption, using lower gains, estimating the state and computing the control law less often (CPU power consumption has been observed to represent a significant fraction of the whole power consumption of the robot Toro~\cite{Henze2019PowerConsumption}).

The proposed analysis doesn't consider maintaining balance by actively using angular momentum (whirling limbs in the air) or modifying the support polygon by making a step. Investigating how uncertainties relate to the decision to make steps, when, how and where, is our next goal.

\section*{Appendix}

If each real coefficient $K(A+BK)^iB$ in the infinite sum~\eqref{Amplification}
is negative, we actually have
\begin{equation}
r=-Kh+1,
\end{equation}
where
\begin{equation}
h=\sum_{i=0}^{\infty}(A+BK)^{i}B.
\end{equation}
By construction, this vector $h$ is the solution of
\begin{equation}
h=(A+BK)\,h+B,
\end{equation}
which can be easily obtained:
\begin{equation}
h=\begin{bmatrix}\frac{1}{1-k}\\0\end{bmatrix},
\end{equation}
resulting in a ratio
\begin{equation}
r=\frac{1}{k-1}+2
\end{equation}
independent from $\lambda$, $\omega$ and $\tau$.

In order to show that this is the case in the gray area of
Fig.~\ref{StabilityAreaWithLines}, factorize the closed-loop matrix as follows:
\begin{equation}
A+BK=M\begin{bmatrix}q_1&0\\0&q_2\end{bmatrix}M^{-1},
\end{equation}
with an invertible matrix $M$, so that:
\begin{align}
r&=\sum_{i=0}^\infty\left|KM\begin{bmatrix}q_{1}^{i}&0\\0&q_{2}^{i}\end{bmatrix}M^{-1}B\right|+1\nonumber\\
&=\sum_{i=0}^\infty\left|\alpha_1q_1^i+\alpha_2q_2^i\right|+1, \label{infSum}
\end{align}
with coefficients $\alpha_1$ and $\alpha_2$ obtained directly from the matrices
$KM$ and $M^{-1}B$. Reorganize each of these terms:
\begin{equation} \label{reorg}
\alpha_1q_1^i+\alpha_2q_2^i=(\alpha_1+\alpha_2)q_1^i+\alpha_2(q_2^i-q_1^i),
\end{equation}
considering that the two poles are positive real and ordered as follows:
\begin{equation}
0\leq q_1\leq q_2<1.
\end{equation}
The first element is negative since we can observe from~\eqref{Amplification}
and~\eqref{infSum}, and then from~\eqref{Trace} that
\begin{align}
\alpha_1+\alpha_2&=KB\\
&=k-k\cosh(\omega\tau)-k\lambda\omega\sinh(\omega\tau)\\
&=q_1+q_2-2\cosh(\omega\tau)<0.
\end{align}

With the help of a computer algebra system, we can actually obtain that
\begin{align}
\alpha_1&=\frac{1-q_1}{(k-1)(q_1-q_2)}(q_1q_2-1+k(1-q_1))\label{eq. Alpha1},\\
\alpha_2&=\frac{1-q_2}{(k-1)(q_2-q_1)}(q_1q_2-1+k(1-q_2))\label{eq. Alpha2}.
\end{align}
Having $\alpha_2$ also negative would complete the proof. The fraction on the
left is positive, so $\alpha_2$ has the same sign as the term on the right. When
$\lambda\geq\omega^{-1}$, this term can be reformulated,
using~\eqref{Determinant}, as
\begin{equation}
k(\cosh(\omega\tau)-\lambda\omega\sinh(\omega\tau)-q_2)\leq k\left(\mathrm{e}^{-\omega\tau}-q_2\right).
\end{equation}
When $\lambda\leq\omega^{-1}$, the gray area satisfies
$k\leq1+\mathrm{e}^{-\omega\tau}$, so
\begin{align}
q_1q_2-1+k(1-q_2)
&\leq q_2^2-1+k(1-q_2)\\
&\leq(1-q_2)(k-1-q_2)\\
&\leq(1-q_2)(\mathrm{e}^{-\omega\tau}-q_2).
\end{align}
In both cases, this term is negative since at least one pole is greater or equal
to $\mathrm{e}^{-\omega\tau}$ in the gray area, so
$q_2\geq\mathrm{e}^{-\omega\tau}$.

%\addtolength{\textheight}{-16cm}

\bibliographystyle{abbrv}
\bibliography{Biblioteca}

\end{document}